\documentclass[a4paper]{article}

\usepackage{INTERSPEECH2020}
\usepackage{multirow}
\usepackage{booktabs}
\usepackage{tikz}

\newcommand\blfootnote[1]{%
  \begingroup
  \renewcommand\thefootnote{}\footnote{#1}%
  \addtocounter{footnote}{-1}%
  \endgroup
}

\title{Automatic Speech Recognition Benchmark for Air-Traffic Communications}
\name{Juan Zuluaga-Gomez$^{1,2}$, Petr Motlicek$^1$, Qingran Zhan$^{1,3}$, Karel Vesely$^4$, Rudolf Braun$^{1}$}
\address{   
  $^1$Idiap Research Institute, Martigny, Switzerland\\
  $^2$Ecole Polytechnique Federale de Lausanne (EPFL), Switzerland\\
  $^3$School of Information and Electronics, Beijing Institute of Technology, Beijing, China\\
  $^4$Brno University of Technology Speech@FIT and IT4I Center of Excellence, Brno, Czechia
  }
\email{\{juan-pablo.zuluaga,petr.motlicek,qzhan,rudolf.braun\}@idiap.ch, iveselyk@fit.vutbr.cz}

\begin{document}

\maketitle
\begin{abstract}	
Advances in Automatic Speech Recognition (ASR) over the last decade opened new areas of speech-based automation such as in Air-Traffic Control (ATC) environments. Currently, voice communication and data links communications are the only way of contact between pilots and Air-Traffic Controllers (ATCo), where the former is the most widely used and the latter is a non-spoken method mandatory for oceanic messages and limited for some domestic issues. ASR systems on ATCo environments inherit increasing complexity due to accents from non-English speakers, cockpit noise, speaker-dependent biases and small in-domain ATC databases for training. Hereby, we introduce CleanSky EC-H2020 ATCO2, a project that aims to develop an ASR-based platform to collect, organize and automatically pre-process ATCo speech-data from air space. This paper conveys an exploratory benchmark of several state-of-the-art ASR models trained on more than 170 hours of ATCo speech-data. We demonstrate that the cross-accent flaws due to speakers' accents are minimized due to the amount of data, making the system feasible for ATC environments. The developed ASR system achieves an averaged word error rate (WER) of 7.75\% across four databases. An additional 35\% relative improvement in WER is achieved on one test set when training a TDNNF system with byte-pair encoding. 
\end{abstract}


\section{Introduction}
\blfootnote{The work was supported by by European Union’s Horizon 2020 project No. 864702 - ATCO2 (Automatic collection and processing of voice data from air-traffic communications), which is a part of Clean Sky Joint Undertaking. The work was also partially supported by SESAR EC project HAAWAII (Highly automated air-traffic controller workstations with artificial intelligence integration). Karel Vesely was also supported by Czech National Science Foundation (GACR) project ``NEUREM3" No. 19-26934X.}
The communication methods between pilots and Air-Traffic Controllers (ATCos) have remained almost unchanged for many decades, where the ATCo's main task is to transfer spoken guidance to pilots during all flight phases (e.g. approach, landing or taxi) and at the same time providing safety, reliability and efficiency. This task has shown to be extremely stressful and highly voice demanding because of the impact a small mistake can make. Several attempts towards increasing the confidence and reducing the workload of pilot-controller communication have been pursued in the past, including experiments with Automatic Speech Recognition (ASR). Initially, due to budget and scarcity of computing power, previous work targeted isolated word recognition, or `voice activity detection' but currently most of the works performs ASR on nearly real time. Military applications were one of the first attempts involving engines for command-related ASR; Beek et al.~\cite{beek1977} contrast the main ASR techniques with its relevance to military applications such as speaker verification, commands recognition and system control of aircraft. They remarked that pilot-ATCo communications have a very limited word set -vocabulary-, speaker-dependent issues and environmental noises that need to be addressed to produce a sufficiently-reliable system. Initially, the integration of ASR technologies in ATCo started in the late 80s' with Hamel et al. report~\cite{hamel1989}; but lately, ASR technologies has been successfully deployed on ATC training simulators. For example, Matrouf et al.~\cite{matrouf1990adapting} proposed a user-friendly and robust system to train ATCos based on hierarchical frames and history of dialogues -context-dependent-. Similarly, DLR~\cite{matrouf1990adapting}, MITRE~\cite{tarakan2008automated} and more recently UPM-AENA~\cite{ferreiros2012speech} under the INVOCA project proposed akin training systems. 

One of the current limitations in developing highly-accurate ASR engines for ATCo communications is the lack of available databases; likewise, generate the transcriptions of such data is extremely costly. Typically a raw ATCo-pilot voice communication recording of one hour -including silences- requires between eight to ten man-hours of transcription effort~\cite{cordero2012automated} (mainly as it requires highly trained participants, often active or retired ATCos). Afterwards, usually only 10 to 15 minutes speech segments of ATCo is obtained from 1h recording (after removing silence segments). Hence, it would take approximately one man-week work to get an hour of ATCos without silences~\cite{ferreiros2012speech, cordero2012automated}.  

Currently, several researchers~\cite{holone2015possibilities} and the International Civil Aviation Organization (ICAO) determined that the air-traffic is expected to grow about 3 to 6 percent yearly at least until 2025. Consequently, it has been seen a huge investment of the European Union (EU) to address the ATCos workload and development of ASR engines for field pilot-ATCos communication, but not only for training purposes. Three recent projects financed by the EU on the scope of ASR for ATCo communications are MALORCA\footnote{MAchine Learning Of speech Recognition models for Controller Assistance, http://www.malorca-project.de/wp/}, ATCO2\footnote{AuTomatic COllection and processing of voice data from Air-Traffic COmmunications, https://www.atco2.org/} and HAAWAII\footnote{Highly Automated Air traffic controller Workstations with Artificial Intelligence Integration}. MALORCA project (together with AcListant\footnote{Active Listening Assistant, www.AcListant.de}) demonstrated that ASR tools can reduce ATCos workload~\cite{helmke2016reducing} and increase the efficiency~\cite{helmke2017increasing}, where also it addressed the lack of transcribed air traffic speech data using semi-supervised training (similar to other ASR related tasks such as speech recognition applied for under-resourced languages~\cite{Imseng_ICASSP_2014,Khonglah_ICASSP2020_2020}) to decrease Word Error Rates (WER) and command error rates~\cite{kleinert2018semi,srinivasamurthy2017semi}. We set as baseline word error rates the results from~\cite{kleinert2018semi,srinivasamurthy2017semi} for two proposed train/test sets. ATCO2 ongoing project aims at developing a unique platform to collect, organize and pre-process air-traffic speech data from air space available either directly through publicly accessible radio frequency channels (such as LiveATC~\cite{liveatc2020}), or indirectly from ANSPs. One of the current challenges of ASR engines for ATCo communications is the changing ATCos accent and vocabularies across different airports; hence, ATCO2 will develop a robust methodology capable of minimize their impact on the system. In this work, we present the first results based on six ATC in-domain databases which, to the authors' knowledge, is the first time that such quantity of command-related databases (spanning more than 170 hours) have been used during the training phase. First, we explore transfer learning from a Deep Neural Network (DNN) system trained on an Out-Of-Domain (OOD) corpus, then we contrast the results with the state-of-the-art ASR chain recipes (from Kaldi's toolkit~\cite{povey2011kaldi}) such as TDNNF and CNN+TDNNF. Secondly, given the high likelihood of Out-Of-Vocabulary (OOV) words due to the intrinsic changing behaviour of the air-space, we tested Byte-Pair Encoding (BPE) based system, which stands as a technique capable of recognise OOV words.

Even though obtaining a full ATCo-pilot communication system goes far beyond of only ASR tasks, we convey in this study a benchmark of experiments going from transfer learning (from an OOD corpus) and adaptation with partial or complete in-domain command-related databases to BPE algorithms and end-to-end TDNNF models. Additionally, this study does not plan to describe in detail intrinsic characteristics of the ATCos communication such as noise (e.g. cockpit noise), speaker's accent, ontology or transcription procedure; but rather, we show technical details and results about ASR in the ATM area. In Section 2 we define the corpus and data preparation used in the proposed benchmark experiments. Section 3 reviews the experimental setup such as lexicon, language and acoustic modelling. Then, Section 4 presents the main results. Finally, Section 5 concludes the paper and proposes the roadmap that ASR systems for ATCo communications should head.

\section{Data Preparation}

Diverse studies conclude that almost 80\% of all pilot radio messages contain at least one error and 30\% of the incidents are accounted by miss-communications (and up to 50\% in the terminal manoeuvring area)~\cite{geacuar2010reducing}. Therefore, ASR stands as a adequate solution, which cannot afford to be trained and tested `on-the-fly' in real operational environment, but it is required to build the best possible system before its deployment. With this intention, we use the state-of-the-art ASR engines that are based on deep neural networks like Time-Delay Neural Networks (TDNN) and Convolutional Neural Network (CNN). These models are known as `data-hungry' algorithms; in fact, state-of-the-art ASR systems need to be trained on large amount of data to achieve and acceptable operational performance. Sadly, there is a lack of such databases in the ATM world. One of our main contribution is to solve this problem employing partly-in-domain or `command-related' databases, which retain similar phraseology and structure.

\subsection{Command-related databases}

One concern that has delayed the development of a unified ASR framework for ATM globally --or at least at country level-- is the vast accent's variability between ATCos from non-English speaking countries. Often, ATCos working in the same country but at different airports may have different accents (e.g. Switzerland). There is also a large variability in dictionary used across airports, as different call-signs, commands, or parameters (e.g. waypoints) are present. Therefore, an un-adapted ASR system will provide significantly worse performance due to unseen accents, OOV words, different recording procedures, parameters or even ontology's variability across different air navigation service providers (ANSPs). Table~\ref{tab:databases} conveys six databases that posses close similarities to ATCo's speech data, accounting to nearly 180 hours (train and test sets). The phraseology and vocabularies are similar across the databases but the speakers' accent is domain-dependent. We also measured the impact of transfer learning of DNN models trained on out-of-domain databases (i.e. Librispeech and Commonvoice, see Table~\ref{tab:databases}) as part of the proposed benchmark. Another pilot-ATCos communication concern are the errors due to OOV words and phonetic di-similarities (e.g. ``hold in position`` and ``holding position", or, ``climb to two thousand" and ``climb two two thousand"). Hence, the ICAO has created a standard phraseology to reduce these errors during the communications; Helmke et al.~\cite{helmke2018ontology} propose a new ontology to transcribe these ATCo-pilots communications.

\begin{table}[t!]
  \caption{Out-of-domain and command-related databases used for transfer-learning (pre-training) and adaptation of TDNNF and CNN+TDNNF models.}
  \label{tab:databases}
  \centering
  \begin{tabular}{ l c p{2.5cm} c }
    \toprule
    	\multicolumn{4}{c}{\textbf{Command-related databases}} \\
    \midrule
     	\multicolumn{1}{c}{\textbf{Database}} & \multicolumn{1}{c}{\textbf{Hours}} & 
     	\multicolumn{1}{c}{\textbf{Accents}} & \multicolumn{1}{c}{\textbf{Ref}} \\
    \midrule
	MALORCA		& 13 & German and Czech & \cite{kleinert2018semi,srinivasamurthy2017semi} \\
	LDC ATCC	& 72.5 & American English & \cite{LDC_ATCC} \\
	HIWIRE		& 28.3 & \raggedright French, Greek, Italian and Spanish & \cite{HIWIRE} \\
	ATCOSIM		& 10.67 & \raggedright German, Swiss German \& French & \cite{ATCOSIM} \\
	UWB ATCC    & 20.6 & Czech & \cite{PILSEN_ATC} \\
	AIRBUS		& 45 & French & \cite{AIRBUS} \\
	\midrule
    	\multicolumn{4}{c}{\textbf{Out-of-domain databases}} \\
    \midrule	
	Librispeech	& 960 & Diverse English & \cite{panayotov2015librispeech} \\
	Commonvoice	& 500 & English subset & \cite{ardila2019common} \\	
    \bottomrule
  \end{tabular}
\end{table}

\subsection{Out-of-domain databases}

As part of the proposed benchmark, we measured the impact of transfer learning to address the lack of in-domain databases. The idea is to pre-train models with well-known out-of-domain databases such as Librispeech~\cite{panayotov2015librispeech} (960 hours) and Commonvoice~\cite{ardila2019common} (500 hours English subset) and then adapt the pre-trained models using in-domain data. The final out-of-domain train set contains nearly 1500 hours of speech data (see Table~\ref{tab:databases}).

\begin{table}[t!]
  \caption{ATC in-domain training and test sets.}
  \label{tab:split_db}
  \centering
  \begin{tabular}{ p{1.1cm} c p{4cm}}
  \toprule
    	\multicolumn{3}{c}{\textbf{Train data-sets}} \\
    \midrule
     	\multicolumn{1}{c}{\textbf{Name}} & \multicolumn{1}{c}{\textbf{Hours}} & 
     	\multicolumn{1}{c}{\textbf{Description}} \\
    \midrule	
	Train1	& 38.7 	& Atcosim (train) + Malorca (Vienna+Prague) + UWB ATCC \\
	Train2	& 137.7	& Airbus + ATCC USA + Hiwire \\
	Tr1+Tr2	& 176.4 & Train1 + Train2 \\
	OOD set & $\sim$1500 & Out-of-domain set: Librispeech + Commonvoice \\
    \toprule
    	\multicolumn{3}{c}{\textbf{Test data-sets}} \\
    \midrule
	Atcosim		& 2.5 	& 20\% of Atcosim train set \\
	Prague		& 2.2 	& From Malorca set \\
	Vienna		& 1.9	& From Malorca set \\
	Airbus 		& 1		& From Airbus set \\
    \bottomrule
  \end{tabular}
\end{table}

\subsection{Databases split}

In order to measure the impact of amount of data, we merged six command-related databases in three training sets as shown in Table~\ref{tab:split_db}. In case of ATCOSIM, we split the database (by speakers) in a 80/20 ratio (i.e. we used 80\% of data as train/validation and the remaining 20\% as test set). In case of MALORCA database, it comprises two ATC approaches (collected from two ANSPs), Vienna and Prague. The remaining databases were collected, processed and released from different projects; we redirect the reader to their references. In fact, Airbus held in 2018 a challenge~\cite{AIRBUS} related to ASR and callsign detection (CSD) of ATCos speech segments; Pellegrini et al.~\cite{pellegrini2019airbus} convey the results of the top 5 teams. It is important to mention that we don't compare our results with theirs because the evaluation set (5 hours) was not released by Airbus; nevertheless, we created from the train data a test set of 350 utterances (1 hour of speech data). The proposed acoustic models are evaluated on four different test sets, where features such as ATCo accent, spoken commands, airport's origin and quantity of training data vary. There are no specific reasons behind the split methodology and the datasets sizes from Table~\ref{tab:split_db}.

\section{Experimental Setup}

\subsection{Lexicon}
The word-list for lexicon was assembled from the transcripts of all the ATCo audio databases (i.e. Tr1+Tr2, see Table \ref{tab:split_db}) and from some other publicly available resources (i.e. lists with names of airlines, airports, ICAO alphabet, etc.). The pronunciations were synthesized with Phonetisaurus~\cite{phonetisaurus}. The G2P (grapheme-to-phoneme) model was trained on Librispeech lexicon, and we inherited its set of phonemes. Likewise, the `spelled' acronyms were auto-detected, and we create their pronunciations separately.

\subsection{Language Modelling}
We train N-gram language models using SRI-LM~\cite{srilm} on the transcripts of the training set Tr1+Tr2 (see Table~\ref{tab:split_db}). We use a tri-gram for the initial decoding and a four-gram model for re-scoring. In our results (Table~\ref{tab:results}) `LM-3' stands for the tri-gram and `LM-4' for the four-gram model. We additionally trained a six-gram LM identified as `LM-6', only for the BPE setup.


\subsection{Experimental Setup}

All experiments are conducted using the Kaldi speech recognition toolkit~\cite{povey2011kaldi}. We report results on two state--of-the-art DNN-based acoustic architectures. We train Factorized TDNN or TDNNF~\cite{povey2018semi} with $\sim$1500 hours of OOD speech (see Table~\ref{tab:databases}) and then we adapt the resulting model with three ATC command-related data-sets (see Subsection~\ref{subs:conv}). Afterwards, we perform flat-start CNN+TDNNF training without any kind of transfer learning or adaptation; the idea behind this is to measure quantitatively whether the amount/accent of training data helps to reduce WERs. We use the standard chain LF-MMI based Kaldi's recipe for both architectures, which includes 3-fold speed perturbation and one third frame sub-sampling. 

\subsection{Conventional LF-MMI Training}
\label{subs:conv}

Conventional LF-MMI training of TDNNF models still relies on a HMM-GMM model to build both the alignments and lattices needed during training. The HMM-GMM models are trained with only the out-of-domain databases i.e. Librispeech + Commonvoice. We followed the standard Kaldi's recipe which requires  100-dimensional i-vector features, 3-fold speed perturbation, and lattices for LF-MMI training supervision, obtained from the training sets. The TDNNF system trained on the out-of-domain training set ($\sim$1500 hours) is tagged as `TDNNF-B'. To measure the impact of the amount of training data on performance in the target domain, we train once with and once without transfer learning on the three different ATC train sets presented in Table~\ref{tab:split_db}. Models trained with transfer learning have `TF' in the name (e.g. TDNN-TF-B). The systems without transfer learning simply are denoted according to their architectures (i.e. TDNNF, CNN+TDNNF or TDNNF-BPE).

\subsection{Byte-Pair Encoding}
\label{subs:bpe}

As part of the benchmark experiments, we use Byte-Pair Encoding (BPE)~\cite{SennrichBPE} on the training transcripts to create a (subword) vocabulary to use for language and acoustic modeling. BPE is a compression algorithm which transforms whole words into `units' of sub-strings, allowing the representation of an open vocabulary where new words can be easily introduced in the lexicons and LMs. There have been several studies using BPE for ASR systems~\cite{BPEDrexler2019, BPEZeyer2018, BPEZeng2019}, we believe there is an especially strong case for ATC communications, as it relies mostly on simple commands and call-signs, but at the same time contains a relatively high amount of foreign proper nouns, which could be missing in a word-based model. For BPE training we limited the number of merges (sub-words) to 2000. We used the original implementation from~\cite{SennrichBPE}. We use a character-based sub-word lexicon which means to get a pronunciation for a word we split the word up into its characters, and then use the most common characters instead of phones.

\begin{table*}[th!]
  \caption{DNN benchmarks with different training methodologies and amount of in-domain and out-of-domain training data. TDNNF-B is our proposed base model trained on Librispeech and Commonvoice. TDNNF-TF-B uses TDNNF-B for initialization (acting as transfer learned model) and then is adapted on the corresponding dataset seen in the table. TDNNF-BPE is a byte-pair encoding system based on 2k sub-word units and a 6-gram LM developed using Tr1+Tr2 train set. CNN+TDNNF is composed of six convolutional layers coupled with nine TDNNF layers at the top (trained on the displayed dataset on the same row).}
  \label{tab:results}
  \centering
  \begin{tabular}{p{1.9cm} c c c c c c c c c c}
    \toprule
        \multirow{3}{*}{\textbf{System}} & \multirow{3}{*}{\textbf{Train Set}} & \multirow{3}{*}{\textbf{Params}} & 
        \multicolumn{8}{c}{\textbf{Word Error Rates (WER) \% - (test sets)}} \\
    	\cline{4-11}
    	&&& \multicolumn{2}{c}{\textbf{Vienna}} & \multicolumn{2}{c}{\textbf{Prague}}
    	& \multicolumn{2}{c}{\textbf{Airbus}} & \multicolumn{2}{c}{\textbf{Atcosim}} \\
    	&&& LM-3 & LM-4 &  LM-3 & LM-4 &  LM-3 & LM-4 &  LM-3 & LM-4 \\
    \midrule
    TDNNF-B & OOD set & 23.1M & 95.8 & 95.8 & 47.6 & 43.3 & 80.6 & 77.5 & 67.5 & 63.4 \\ 
    \midrule
	\multirow{3}{1in}{TDNNF-TF-B} & Train1 & 
	\multirow{3}{*}{20.8M} & 7.6 & 7.1 & 9.1 & 9.0 & 53.6 & 51.4 & 7.5 & 7.3\\
	& Train2 && 30.2 & 26.2 & 19.3 & 17.8 & 14.9 & 14.6 & 23.9 & 20.5\\
	& Tr1+Tr2 && 7.5 & 6.9 & 8.6 & 8.4 & 15.2 & 14.7 & 5.9 & 6.0\\
	\midrule
	\multirow{3}{1in}{TDNNF} & Train1 & 
	\multirow{3}{*}{20.8M} & 8.1 & 7.5 & 8.9 & 8.7 & 67.8 & 66.7 & 8.5 & 8.1\\
	& Train2 && 33.2 & 30.2 & 20.1 & 18.8 & 14.6 & 14.5 & 23.4 & 19.6\\
	& Tr1+Tr2 && \textbf{7.1} & \textbf{6.6} & 8.1 & 7.9 & \textbf{14.6} & 
	\textbf{14.4} & 5.3 & 5.2 \\ 
	\midrule
    CNN+TDNNF & Tr1+Tr2 & 14.3M & 7.1 & 6.7 & 8.1 & 7.9 & 15.1 & 14.7 &
    \textbf{5.0} & \textbf{5.1} \\
    &&& \multicolumn{2}{c}{LM-6} & \multicolumn{2}{c}{LM-6} & 
    \multicolumn{2}{c}{LM-6} & \multicolumn{2}{c}{LM-6} \\
    \cline{4-11}
    TDNNF-BPE & Tr1+Tr2 & 20.8M & \multicolumn{2}{c}{7.6} & \multicolumn{2}{c}{\textbf{5.1}} &
    \multicolumn{2}{c}{15.1} & \multicolumn{2}{c}{7.2} \\
    \bottomrule
  \end{tabular}
\end{table*}

\section{Results and Discussion}
\label{sec:results}

The results (Table~\ref{tab:results}) are split into four blocks. First, TDNNF-B is trained on an OOD set consisting of 1500 hours. This is our base model to perform transfer learning. Second, TDNNF-B model is adapted to the different ATC datasets (using TDNNF-B as initialization) i.e. Train1, Train2 and Tr1+Tr2. Third, we compare WERs for TDNNFs without transfer learning. Finally, we present results on a CNN+TDNNF chain model and TDNNF trained with BPE units. We kept the same hyper-parameters across all the experiments in order to make fair comparisons between models. The base model performs poorly on the ATC data. This is not surprising as Librispeech and Commonvoice are both read speech with mostly clear audio. The ATC data is more noisy, the speakers talk much quicker, and the accents are stronger. Despite the significant difference in domains, the pretraining still helps when the target dataset is not too large, as can be seen when comparing the first two rows of the TDNNF-TF-B and the TDNNF models in Table~\ref{tab:results} (trained on Train1 or Train2). Note the sometimes large differences in performance of the models trained on either Train1 or Train2 can be explained by whether the accent(s) in the test set were also present in the training set. Once the target domain dataset becomes large enough, we do not see the benefit of pretraining (see the last row of the TDNNF-TF-B and the TDNNF models). The last block of experiments provides a broader cover of different DNN architectures and techniques on our proposed ASR benchmark for ATC communications. There is no clear winner. The CNN+TDNNF system yielded a new baseline of 5\% WER for Atcosim, showing a relative improvement on WERs of 16.7\% and 3.9\% when compared to TDNNF-TF-B and TDNNF. The best model for Vienna dataset was TDNNF trained on Tr1+Tr2 and scored with a 4-gram LM, whereas for Prague it was TDNNF with 6-gram and lexicon based on BPE. Compared to previous experiments on MALORCA~\cite{kleinert2018semi,srinivasamurthy2017semi}, our approach yields 29.8\% and 37.9\% relative WER improvement for Vienna and Prague. 

We further investigated why the BPE model does significantly better on the Prague test set, and found that the difference in performance is entirely explained by reduced deletions (five times more deletions in TDNNF and CNN+TDNNF than TDNNF-BPE system). The word-based model is obviously not able to recognize OOV words, which is the primary reason for the deletion errors. The OOV rates on Prague, Vienna, Airbus and Atcosim test sets are 3.3\%, 1.1\%, 0.0\% and 0.1\%. This shows that the BPE system is capable of recognizing OOVs and thereby improving performance; although, it does come at a cost since the BPE models also perform significantly worse on some test sets. Further investigation is required to understand the differences in performance between word and sub-word (BPE) based systems. We noticed that the BPE based model performs better on foreign words (even when the word-based model includes these words in its lexicon).  We attributed it to the character-based lexicon system, which generalizes better to foreign languages which are not closely related to English. The Atcosim baseline WER is presented in~\cite{holone2016n}, schiving 8.5\% absolute WER when performing n-best list re-ranking using syntactic knowledge. In our case, we obtain first 63.4\% WER with TDNNF-B and an improvement to 8.1\% absolute WER when training only on Train1 set. An additional 10\% relative WER improvement can be obtained if employing transfer learning (i.e. TDNNF-TF-B + Train1), reaching 7.3\% absolute WER. An additional 28\% relative WER improvement is achieved when training TDNNF on Tr1+Tr2, which is a bigger train set. Finally, with the intention to explore different DNN architectures we were able to further reduce the absolute WER to 5.0\% when using a CNN+TDNNF system trained on Tr1+Tr2, accounting to 3.8\% relative improvement from TDNNF. 

Finally, we are currently exploring new hybrid frameworks for acoustic modeling --of ATCo data-- such as Pkwrap~\cite{pkwrap20}. Pkwrap~\footnote{https://github.com/idiap/pkwrap} is a simple wrapper that is useful to train acoustic models in PyTorch using Kaldi's LF-MMI training framework.


\section{Conclusions}

This work introduce state-of-the-art DNN architectures to the area of ASR for air-traffic communications. We tested several DNN architectures, amount of training data and transfer learning across the given experiments in order to reasonably compare their performance. To the author's knowledge, this is the first study employing six air-traffic command-related databases spanning more than 176 hours of speech data that are strongly related in both, phraseology and structure to ATCos-pilots communications, therefore dealing with the burden of lack of databases that many previous studies have quoted. Specifically, we have shown that using in-domain ATC databases, even if not from the same country/airport, the system is capable to yield a 29.8\% and 37.9\% relative WER improvement for Vienna and Prague approaches. Also, we reported new baselines for Vienna, Prague and Atcosim test sets. In general, our proposed ASR engine for ATCo speech achieves an averaged WER of 7.75\% across four test sets with different accent; the given results suggest that systems trained on cross-accent data helps in the overall system's performance, rather than limiting the amount of data to single-accent datasets. Finally, one of the main outcomes of this research was the results on byte-pair encoding with Prague approach, reaching 5.0\% WER. We advise that future research should further explore BPE units for ATC. 

\bibliographystyle{IEEEtran}
\bibliography{main}

\end{document}